# A Radar-Shaped Statistic for Testing and Visualizing Uniformity Properties in Computer Experiments

Jessica Franco[1], Laurent Carraro[1], Olivier Roustant[1] and Astrid Jourdan[2].

[1]Département 3MI Ecole Nationale Supérieure des Mines, Saint-Etienne, nom@emse.fr.
[2]Ecole Internationale des Sciences du Traitement de l'Information, Pau, astrid.jourdan@eisti.fr.

In the study of computer codes, filling space as uniformly as possible is important to describe the complexity of the investigated phenomenon. However, this property is not conserved by reducing the dimension. Some numeric experiment designs are conceived in this sense as Latin hypercubes or orthogonal arrays, but they consider only the projections onto the axes or the coordinate planes. In this article we introduce a statistic which allows studying the good distribution of points according to all 1-dimensional projections. By angularly scanning the domain, we obtain a radar type representation, allowing the uniformity defects of a design to be identified with respect to its projections onto straight lines. The advantages of this new tool are demonstrated on usual examples of space-filling designs (SFD) and a global statistic independent of the angle of rotation is studied.

## 1. INTRODUCTION

For the last 15 years or so, the design of experiment theory initiated by Fisher (1926) has experienced a revival for the analysis of costly industrial computer codes. This development has led to at least two major changes. First, these codes represent phenomena of an increasing complexity, which implies that the corresponding models are often nonlinear and/or nonparametric. Second, the experiment itself is different. Numerical experiments are simulations and, except for stochastic codes implementing a Monte Carlo-based method, produce the same response for identical conditions (including algorithm and computer-based parameters). Therefore, repeating an experiment under the same conditions does not make sense since no new information is acquired.

In this new paradigm, the experiment planning methods are different. For example when the code is to be analyzed for the first time before any simulation has been made (scanning phase), one often tries to satisfy the following two requirements. Firstly, distribute the points in the space as uniformly as possible to catch non-linearities; this excludes repetitions also. Secondly, this space coverage should remain well-distributed even when the effective dimension is lowered. The first requirement was the starting point of research work in space filling designs (SFD). The quality of the spatial distribution is measured either by using deterministic criteria like minimax or maximin distances (Johnson, Moore and Ylvisaker, 1990), or by using statistical criteria like discrepancy (Niederreiter, 1987, Hickernell, 1998,

Fang, Li, Sudjianto, 2006). The second requirement stems from the observation that codes often depend only on a few influential variables, which may be either direct factors or "principal components" composed of linear combinations of these variables. Therefore dealing only with these influential factors is more efficient. Note that dimension reducing techniques like SIR (Li, 1991) or KDR (Fukumizu, Bach, Jordan, 2004) effectively identify the space generated by the principal factors. Hence, it is desirable that the space filling property should be also satisfied *in the projection onto subspaces*. This is the aim of Latin hypercubes designs (LHD) and orthogonal arrays (OA) in computer experiments. For instance, the space filling LHDs provide a uniform repartition in the projection onto margins, so that there is no loss of information if the code depends on only one variable. In addition, orthogonal arrays with space-filling properties extend this aspect to higher dimensions (see, for example, Koehler and Owen, 1996, and more generally, Owen, 1992 or Santner, Williams, Notz, 2003). Nevertheless, considering only the projections onto margins is not sufficient if, for example, the code is a function of a linear combination of 2 variables.

In this article we introduce a statistic based on 1-dimensional projections to test the uniformity for a design of experiment. The choice of 1 dimension is due to the difficulty in computing the theoretical distribution for higher dimensional space. The advantage of restricting to a single dimension is that it offers a simple viewing tool similar to a radar screen. By representing the statistic's value in all directions, one obtains a parameterized curve (or surface), identifying the uniformity defects of a design with respect to its projections onto straight lines. The article is structured as follows. In section 2 we define the new statistic and the associated visualization tool, called uniformity radar, and give some properties of them. In section 3 we show examples of radar applications to space-filling designs. Section 4 is devoted to extend the concept by defining a global statistic which does not depend on a particular axis of rotation. In section 5, a discussion on the uniformity radar allows specifying its scope of application. Proofs are given in the appendix.

## 2. UNIFORMITY RADAR

In the analysis of a computer code, let us consider a uniform experiment design on a cubic domain $\Omega = [-1,1]^d$ . Note $x_1,...,x_N$ the experimental points, and $(H_0)$ the hypothesis " $x_1,...,x_N$ were generated by independent random sampling according to the uniform distribution in $\Omega$ ". If the computer code depends only on one principal component, the projections on this axis should be well distributed. More generally, denote $L_a$ the straight line generated by the unitary vector $a = (a_1,...,a_d)$ of $\Omega$, and $\mu_a$ the probability distribution of the projections of $x_1,...,x_N$ onto $L_a$. Ideally, we may expect that in any direction $a$ the distribution $\mu_a$ is uniform. However, this is not realistic when $L_a$ is not a coordinate axis. For example, in the case of the cubic domain $[-1,1]^2$, the density of the projected design points is higher in the central part of the axis as can be seen in Figure 1.

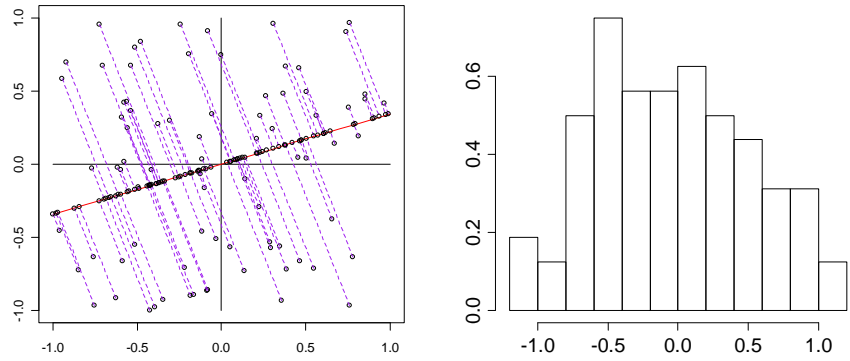

*Figure 1. Left : Projections of points onto an axis $L_a$. Right : The histogram of the projections.*

More precisely, this distribution is supported by 3 areas defined by the projection of the corners of the domain. The distribution $\mu_a$ is continuous, with density represented below. The nodes of the trapezoidal density correspond to the corners of the square $\Omega$ projected onto the axis $L_a$, where $a = (\cos\theta, \sin\theta)$.

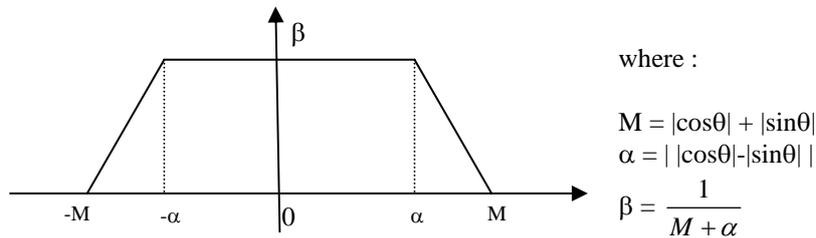

*Figure 2. The distribution of the projections for a 3 dimensional cubic domain.*

In the general case, the projection onto $L_a$ is a linear combination of independent random variables of uniform distribution, which leads to a traditional problem of probabilities first solved by Lagrange in the 18th century (see discussion in Elias and Shiu (1987) on this topic). If we denote $X = (X_1, ..., X_d)$, where $X_1, ..., X_d$ are independent random variables identically distributed following the uniform distribution over $[-1,1]$, and $Z$ the projection of $X$ onto $L_a$, where $a_j \neq 0 \ \forall j \in \{1, ..., d\}$, then the distribution function of $Z$ is given by :

$$F_Z(z) = \left( \prod_{j=1}^{d} \frac{1}{2a_j} \right) \times \sum_{s \in \{-1,1\}^d} \varepsilon(s) \frac{(z + s.a)_+^d}{d!}$$

where $s = (s_1, ..., s_d) \in \{-1,1\}^d$ are the corners of the hypercube $\Omega = [-1,1]^d$, $\varepsilon(s) = \prod_{j=1}^{d} s_j$, $s.a$ is the scalar product of the vectors $s$ and $a$, and $(y)_+ = \max(y, 0)$ the positive part of $y$. As a result, for a given axis, $Z$ admits a piecewise linear density whose nodes correspond to the projections of the corners of the domain.

*Note.* The distribution of the projections is known in other situations, as in the case of a spherical domain: if $\Omega$ is the unit sphere of $R^d$, a direct calculation shows that $\mu_a$ admits the density $f_a(x) = \frac{2}{\pi} \sqrt{1-x^2} 1_{[-1,1]}(x)$, the distribution function being equal to $F_a(x) = \frac{1}{2} + \frac{1}{\pi}(\text{Arc} \sin x + x\sqrt{1-x^2})$ for $x \in [-1,1]$.

In sum, for a uniform experiment design to have good distribution properties on the 1-dimensional projections, it will be necessary that in all the directions $a$, the empirical distribution of the projections onto $L_a$ is close to their theoretical distribution under the hypothesis $(H_0)$. There exist many distribution adequacy statistics (see D'agostino and Stephens, 1986), which allow for a large number of choices to define a criterion adapted for this purpose. However, possibilities are limited by special requirements. To start with, it is preferable that the statistic's distribution be known to avoid the approximate calculation of its distribution. Furthermore, one would also like the statistic to be distribution free, that is, its distribution doesn't depend on the projection direction to have a unique rejection threshold for all the angles. Also, for the sake of consistency, it would be desirable for the retained statistic to be interpretable in terms of discrepancy when projections are made onto a coordinate axis. Finally, two famous statistics (at least) correspond to these requirements: the Kolmogorov-Smirnov statistic

$$D_N(a) = \sup \left| \hat{F}_{N,a}(z) - F_a(z) \right| \tag{1}$$

and the Cramér-Von Mises statistic

$$N\omega_N^{\,2}(a) = \int (\hat{F}_{N,a}(z) - F_a(z))^2 \, dz \tag{2}$$

where $\hat{F}_{N,a}$ is the empirical distribution function of the projections of $x_1,...,x_N$ onto $L_a$, and $F_a$ the distribution function of $\mu_a$. When $L_a$ is a coordinate axis, $\mu_a$ is the uniform distribution on [0,1] and these statistics correspond, respectively, to the discrepancies $L_\infty$ and $L_2$ (Niederreiter, 1987). In what follows, we decided to work on the first because the conclusions seem equivalent, while the corresponding graphics are a little more readable (see section 5). By analogy with the case of coordinate axes, we will talk about a *discrepancy of projections* to designate the Kolmogorov-Smirnov statistic of the formula (1).

The discrepancy of projections provides a tool for visualizing the defects in uniformity based on the 1-dimensional projections. Since this tool looks like a radar screen, we propose to call it *uniformity radar*. Its utilization depends on the dimension of $\Omega$. In 2 dimensions, the discrepancy of projections is calculated in 360° by continuously projecting onto a rotating axis. Thus, one obtains a parameterized curve in polar coordinates

$$\theta \mapsto D_N(\theta)$$

defined over [0,2π], called *2D radar*. By displaying the quality of distributions in all directions, the 2D radar provides decision support on design uniformity. In 3 dimensions, one calculates the discrepancy of projections onto an axis $L_{\theta,\varphi}$, for all directions, pivoting around the center of the domain. This axis is defined in spherical coordinates by an angle $\theta$ in longitude and $\varphi$ in latitude. This time a parameterized surface is obtained, called *3D radar*,

$$(\theta, \varphi) \mapsto D_N(\theta, \varphi)$$

defined over $[0, 2\pi] \times \left[ -\frac{\pi}{2}, \frac{\pi}{2} \right]$. In higher dimensions, it seems unrealistic to make an angular scan of the space $\Omega$. In addition, it becomes almost unfeasible to represent the result graphically (although the calculation is still possible). However, the hypothesis ($H_0$) remains valid on 2- and 3-dimensional coordinate spaces. Therefore, the uniformity radar may be applied to all pairs and/or triplets of possible dimensions.

In practice, the quality of the representation can be degraded by discretizing. Here the 2D and 3D radars are continuous applications. But $D_N$ is not differentiable with respect to all the axes $L_a$ so that at least two points of $\Omega$ are projected onto the same point, which explains why the parameterized curves of the next section are not smooth and contain many singularities.

## 3. APPLICATIONS OF UNIFORMITY RADAR

In this section we present a few examples to show the interest in using uniformity radar to test the uniformity of the distribution of experimental points by looking at their projection. We consider cases where the hypothesis $(H_0)$ of a uniform distribution in the experimental domain is plausible. For each representation of the uniformity radar we have added the circle (or the sphere) of radius *ks* equal to the statistic of the Kolmogorov-Smirnov test associated with a 95% confidence level. Recall that since the statistic is distribution free, *ks* does not depend on *a*. This provides a decision-making support or, at least, a means of comparison with the random designs obtained by a uniform sampling. Should the studied design be stochastic (pseudorandom, Latin hypercubes or randomized orthogonal arrays), the graph displays the directions *a* along which the hypothesis $(H_0)$ is rejected. If the design is deterministic, we are outside the usual scope of application of the test. If one of the values of $D_N(a)$ is greater than *ks*, then we can only say that this design is worse than a random design since the probability that a random design will have a better discrepancy exceeds 95%. The following examples apply essentially to the latter cases because we preferred to use known SFD designs without transforming them. Nonetheless, in practice, it would suffice to apply randomization or scrambling (see, for example, Fang, Li, Sudjianto, 2006) to obtain a stochastic design, and thus be under the usual assumptions of statistical tests.

*Example 1. Analysis of a 15-dimensional Halton sequence using 2D radar.* We consider the first 250 points of a 15-dimensional Halton sequence of low discrepancy (1960). Since the design is high dimensional, we apply the radar to all pairs of possible dimensions. Among the rejected pairs we have, for example, the pair (14, 15), represented on Figure 3.

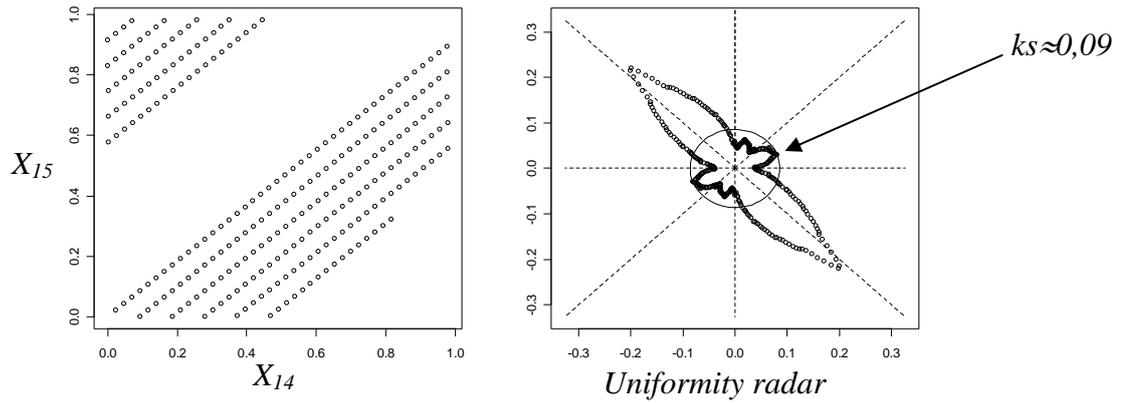

*Figure 3. Left: Design projections of a Halton sequence onto the plane ($X_{14}$,$X_{15}$). Right: The 2D radar curve.*

In this example, since there exist values of $D_N(a)$ outside the circle of radius *ks*, the uniformity radar detects a non-uniform distribution for the 2-dimensional projections onto $(X_{14}, X_{15})$. The largest deviation in uniformity is observed in direction *a* associated with an angle of approximately 135°, corresponding to the direction that is perpendicular to the visible diagonal alignments on the figure to the left. Here, we find ourselves faced with the well-known defect of high-dimensional Halton sequences, which do not preserve a low discrepancy in projection (Thiémard (2000), Morokoff, Caflisch (1994)). Note, however, that the radar is not designed to systematically detect directions of alignment as we will see in the next example.

*Example 2. Analysis of orthogonal arrays using 3D radar.* Let us consider a 49 points linear orthogonal array of strength two in 3 dimensions (Owen (1992) and Jourdan (2000)). Like Latin hypercubes, these designs are often recommended for numeric experimental designs because of their appreciable properties in projection. Projected onto a surface, an orthogonal array of strength two always yields a regular grid of points. However, the non-redundancy of the 2-dimensional projections does not imply a good distribution of the points neither on the axes of the domain (here 7 packets of 7 points) nor in the 3D space, as we will see.

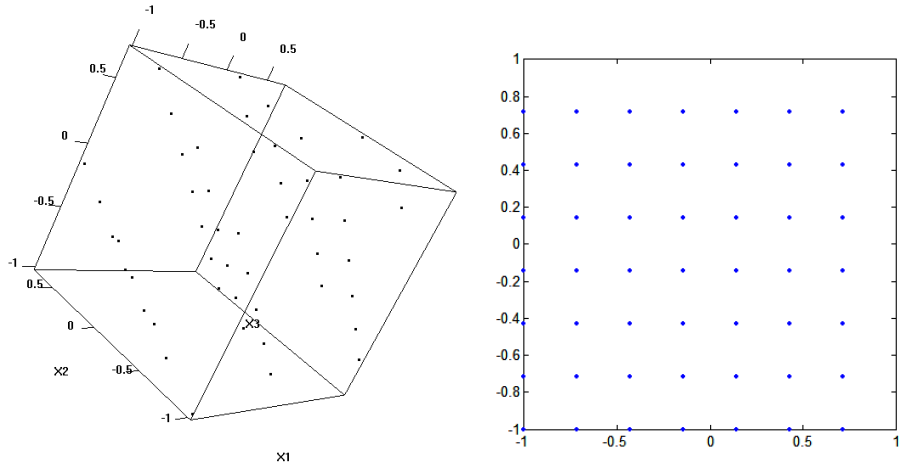

*Figure 4. Left. A 3-dimensional linear orthogonal array of strength two with 49 points. Right. Projections onto the plane ($X_1, X_2$).*

The considered design is a linear orthogonal array of strength two. The way it is built implies that the points satisfy the relation $x_1+3x_2+x_3=0$ (mod 7). Therefore, these points are located on 5 parallel planes. As a result, the distribution of the projections onto the axis perpendicular to these planes will not be satisfactory. This problem is never mentioned in computer experiments. However, it is well-known in the literature of experimental designs.

We applied the uniformity radar to the studied orthogonal array and represented the sphere of radius *ks* (the Kolmogorov-Smirnov statistic with a 95% confidence level), the points $D_N(\theta, \varphi)$, and drew straight lines joining these points to the surface of the sphere to illustrate directionally the deviations in uniformity. We also represented the logarithm of the p-values of the Kolmogorov-Smirnov test as a function of $\theta$ and $\varphi$. We observe that the radar does indeed detect a problem in the direction perpendicular to the 5 planes, $\theta = 72°, \varphi = 18°$. In addition, it reveals a poor distribution of the projected points onto the direction $\theta = 35°, \varphi = 40°$. This is a problem that we could not have anticipated from the design's characteristics.

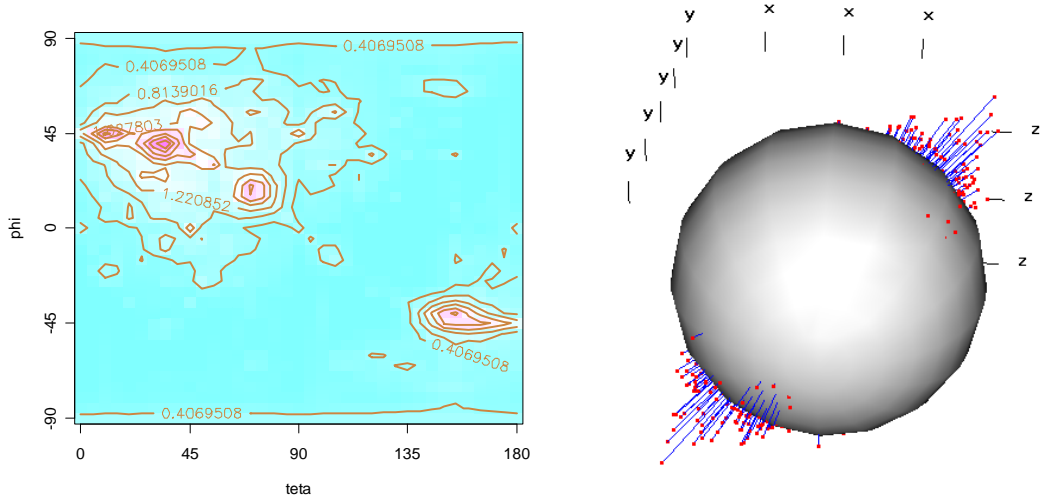

*Figure 5. Left. The p-values in –log10 of the Kolmogorov-Smirnov test. Right. The values of the Kolmogorov Smirnov test with the radar's representation in pins.*

However, the radar does not detect a problem on the coordinate axes, for which the projections are stacked in 7 packets of 7 points. This can be explained by observing the empirical cumulative distribution function (ecdf). For example, the deviation of the (transformed) ecdf of the projected points onto (Oz) from the uniform cdf is not large, as seen in Figure 6(c). Actually, the alignments can be detected by the Kolmogorov-Smirnov statistic, especially when they are not regularly distributed in space (as in example 1). To illustrate this point, we represented the (transformed) ecdf of the projected points onto the axis L1, L2, L3 corresponding to, respectively, the direction $\theta = 35°, \varphi = 40°$, the axis perpendicular to the 5 parallel planes, and (0z). The deviation in uniformity is much larger on figures (a) and (b).

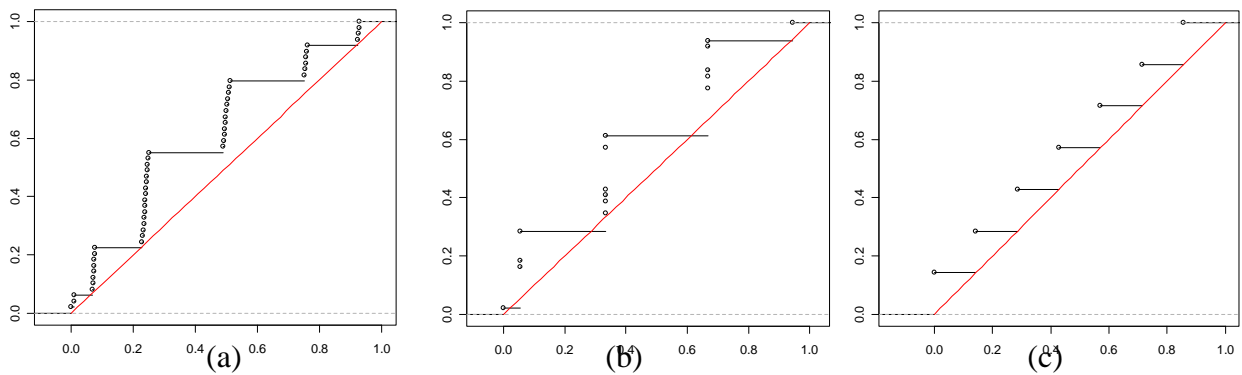

*Figure 6. Left to right. the distribution functions (after transformation) of the projected points onto L1, L2 and L3.*

## 4. A GLOBAL STATISTIC FOR 2D RADAR

*Example 3. Toward an extension of the uniformity radar.* Let us consider the first 100 points of an 8-dimensional Halton sequence projected onto the subspace formed by $(X_3, X_6)$.

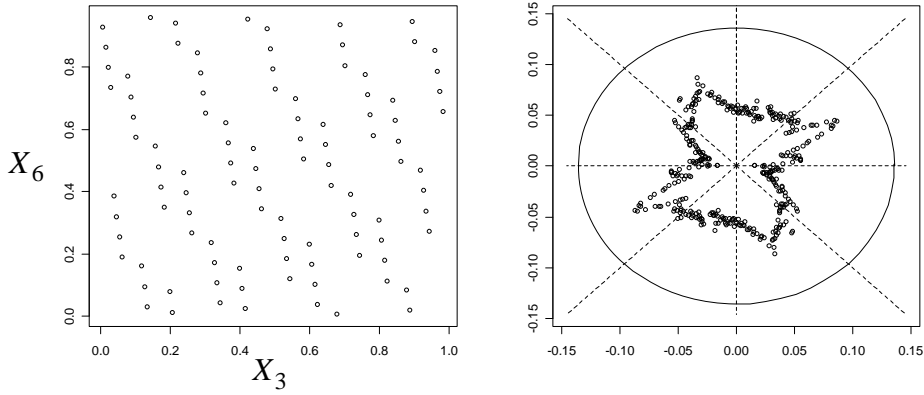

*Figure 7. The first 100 terms of an 8-dimensional Halton sequence projected onto $(X_3, X_6)$.*

Note that all the points of the uniformity radar are inside the circle of radius *ks* and, as a result, the radar accepts the design although we can see that the points of the plane $(X_3, X_6)$ are not uniformly distributed. However, the discrepancy values are rather scattered with a low value for angle $\theta = 0°$, and rather high values, for example, for angle $\theta = 29°$, which seems to correspond to the orthogonal direction of alignments. The idea to reject this type of design amounts, therefore, to defining a new statistic which introduces minimum and maximum discrepancies. In order to avoid scale problems, we suggest taking the ratio of these quantities. We have:

$$G_N = \frac{\sup_{\theta \in [0, 2\pi]} D_N(\theta)}{\inf_{\theta \in [0, 2\pi]} D_N(\theta)} \qquad (3)$$

This statistic should filter out designs which have a relative poor distribution in one direction. This statistic has the advantage of being global. This increase the power of the corresponding statistical test. For a fixed value of N, the distribution of $G_N$ seems difficult to obtain other than by simulation. In the appendix, the table corresponding to N=1…100 is given. According to this table, the rejection threshold at level 95% for a design of 100 points is equal to 4.23. In example 3, the observed value of the statistic $G_N$ is equal to 6.07, which very clearly eliminates this design.

In the statistic table, we observe that the values stabilize as N increases, which leads us to believe that it is possible to specify its asymptotic behavior. This is in fact the case. The following demonstration is based on the theory of empirical processes.

*Asymptotic distribution of the global statistic (3).* Let $X = (X_1,..., X_N)$, where $X_1,..., X_N$ are independent random variables from uniform distribution on $[-1,1]^2$. Denote $R_\theta(X)$ as the projection of $X$ on axis $L_\theta$. Let $\left(Y^{N,\theta}_{A_t}\right)_{A_t}$ be the following process:

$$Y^{N,\theta}_{A_t} = \frac{1}{N}\sum_{i=1}^{N} 1_{]-\infty,t]}(R_\theta(X_i)) - P(R_\theta(X) < t) = \frac{1}{N}\sum_{i=1}^{N} 1_{A_t}(X_i) - P(X \in A_t)$$

where $A_t = \{x \in R^d, R_\theta(x) < t\}$.

Using these notations, the discrepancy of projections may be written as $D_N(\theta) = \sup_{t \in \mathbb{R}} \left|Y^{N,\theta}_{A_t}\right|$.

Therefore,

$$G_N = \frac{\sup_\theta D_N(\theta)}{\inf_\theta D_N(\theta)} = \frac{\sup_\theta \sup_t \left|Y^{N,\theta}_{A_t}\right|}{\inf_\theta \sup_t \left|Y^{N,\theta}_{A_t}\right|}.$$

The limit distribution of the process $\left(Y^{N,\theta}_{A_t}\right)_{A_t}$ is obtained using the theory of empirical processes. With the central limit theorem, one obtains the convergence of the finite dimensional distributions toward the Gaussian process of same expectation and covariances. Furthermore, the family $(A_t)_t$ is a VC-class (see …). Therefore,

$$\left(Y^{N,\theta}_{A_t}\right)_{A_t} \xrightarrow{loi} \left(Y^{\theta}_{A_t}\right)_{A_t}$$

where $\left(Y^{\theta}_{A_t}\right)_{A_t}$ is a centered Gaussian process with covariance:

$$\mathrm{cov}\left(Y^{\mu}_{A_s}, Y^{\theta}_{A_t}\right) = \frac{1}{N^2}\sum_{i,j}^{N} P\left(R_{\mu}(X)<s, R_{\theta}(X)<t\right)$$

$$-\frac{1}{N}\sum_{i}^{N} P\left(R_{\mu}(X)<s\right).P\left(R_{\theta}(X)<t\right)$$

$$-\frac{1}{N}\sum_{j}^{N} P\left(R_{\theta}(X)<t\right).P\left(R_{\mu}(X)<s\right)$$

$$+ P\left(R_{\mu}(X)<s\right).P\left(R_{\theta}(X)<t\right)$$

$$\mathrm{cov}\left(Y^{\mu}_{A_s}, Y^{\theta}_{A_t}\right) = P\left(R_{\mu}(X)<s, R_{\theta}(X)<t\right) - P\left(R_{\mu}(X)<s\right).P\left(R_{\theta}(X)<t\right)$$

$$\mathrm{cov}\left(Y^{\mu}_{A_s}, Y^{\theta}_{A_t}\right) = P(A_s \cap A_t) - P(A_s).P(A_t). \tag{4}$$

and

$$P(G_N < y) \xrightarrow[N \to +\infty]{} P\left(\frac{\sup_{\theta}\sup_{t}\left|Y^{\theta}_{A_t}\right|}{\inf_{\theta}\sup_{t}\left|Y^{\theta}_{A_t}\right|} < y\right)$$

The probability $P(A_s \cap A_t)$ is interpreted as the surface of a polygon delimited by the domain and the straight lines $A_s$ et $A_t$ (see Figure 8). This surface can be analytically calculated, by noting (for example) that the polygon is a partition of triangles with a common corner to be selected from among the vertices of the polygon. The asymptotic distribution of $G_N$ can then be obtained by simulating a centered Gaussian field with covariance matrix defined by (4).

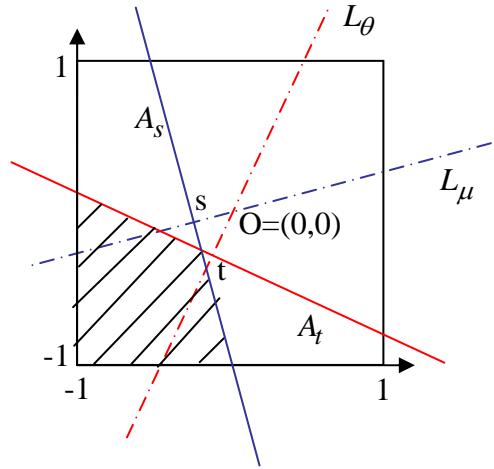

*Figure 8. Interpretation of the probability $P(A_s \cap A_t)$.*

The procedure is semi-analytic because simulations are needed to perform the calculations. In theory, the global statistic may be extended to 3 dimensions. For a finite sample, the statistic table can be calculated with simulations. However, the asymptotic distribution is much more difficult to obtain.

## 5. DISCUSSION

Due to the complexity of the phenomena simulated by computer codes, which implies non-linearities, distributing numerical experiments as uniformly as possible in the domain is preferable. In addition, this distribution should continue to be satisfactory when projecting onto subspaces, especially when the code depends only on a small number of factors or principal components. In this article we introduce a statistic based on 1-dimensional projections to test the uniformity for a design of experiment. In 2 and 3 dimensions, it provides a simple viewing tool called uniformity radar, which graphically tests the uniformity hypothesis by omni-directional scanning. Moreover, we introduced a global statistic in 2 dimensions to further identify unsatisfactory designs that had been accepted by the uniformity radar.

These tools were applied on usual SFD designs. On these case studies, some of the designs have very poor properties when projecting to subspaces, as the 15-dimensional sequence with low discrepancy in example 1 or the 3-dimensional orthogonal array in example 2. The uniformity radar was able to detect these defects. It succeeds when there is a rectangular shaped empty area in the domain, as in the aforementioned examples. In such cases, the distribution is unsatisfactory when projecting onto the rectangle's width. The uniformity radar can also detect the alignments of points, but may not identify them if the directions of alignments are well distributed such as with a factorial design (see example 2). This underscores the lack of power of the Kolmogorov-Smirnov test when the sample is generated from a continuous distribution supported by the union of small intervals regularly distributed. In practice, this situation is not very detrimental because the SFD obtained by a deterministic process are often randomized or scrambled (see, for example, Fang, Li and Sudjianto, 2006).

Our uniformity radar may be adapted to other goodness-of-fit statistics, such as Cramér-Von Mises (see section 2), which corresponds to the discrepancy $L_2$ for a projection onto a coordinate axis. For instance, we repeated the examples 1 to 3 with the corresponding radar. As expected, the conclusions are the same because the Kolmogorov-Smirnov and Cramér-Von Mises tests do not present any clear-cut difference in terms of power. Interestingly, the main difference is graphic. The curve of the radar defined with the Cramér-Von Mises statistic is smoother, which is due to the norm $L_2$, and introduces sometimes large scale variations from one design to another, while these differences are attenuated by the norm $L_\infty$ in the examples given here. For this reason, the $L_\infty$ radar may be preferred because the conclusions are more apparent.

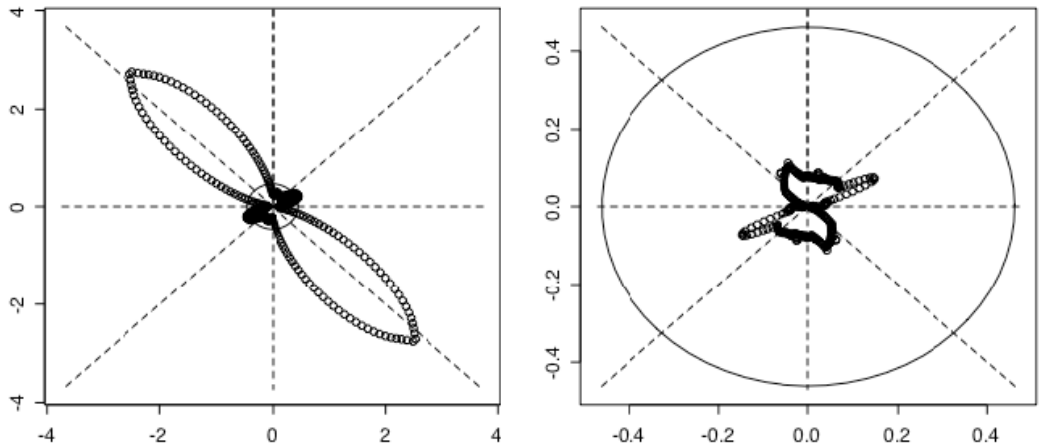

*Figure 9. Uniformity radar with the Cramér-Von Mises statistic for examples 1 and 3.*

ACKNOWLEDGMENTS

This work was conducted within the frame of the DICE (Deep Inside Computer Experiments) Consortium between ARMINES, Renault, EDF, IRSN, ONERA and TOTAL S.A.
We also thank Chris Yukna for his help on editing.

APPENDICES

*Distribution of a linear combination for uniformly distributed independent variables.*

The final result demonstrated by Ostrowski (1952) often uses very technical methods. However, Elias et al. (1987) propose a simpler operational computational method that we apply to our special case.

Let $Z = X.a = X_1 a_1 + ... + X_d a_d$. This is, therefore, a sum of independent random variables of uniform distribution on $[-a_i, a_i], i = 1, ..., d$. As a result, it admits the density given by

$$f_Z = \left(\prod_{j=1}^{d} \frac{1}{2a_j}\right) \times \left(1_{[-a_1, a_1]} * \cdots * 1_{[-a_d, a_d]}\right),$$

where $*$ designates the convolution product. To calculate this product of convolutions, Elias et Shiu (1987) suggest using translations. Let $T_a(f)(x) = f(x + a)$, the translation of the function *f* from *-a*. Then we can write (with the convention $0^\alpha = 0, \forall \alpha \geq 0$):

$$1_{[-a_j, a_j]}(x) = (x + a_j)_+^0 - (x - a_j)_+^0$$
$$= T_{a_j}\left(x_+^0\right)(x) - T_{-a_j}\left(x_+^0\right)(x)$$
$$= \left(T_{a_j} - T_{-a_j}\right)\left(x_+^0\right)(x)$$

And, therefore, $f_Z(x) = \left(\prod_{j=1}^{d} \frac{1}{2a_j}\right) \times \left(\left(T_{a_1} - T_{-a_1}\right)\left(x_+^0\right) * \cdots * \left(T_{a_d} - T_{-a_d}\right)\left(x_+^0\right)\right)$.

The proof follows from two basic lemmas.

*Lemma 1.* The translation commutes with the convolution.

$$\left(T_a(f_{x_1})\right) * f_{x_2} = f_{x_1} * \left(T_a(f_{x_2})\right) = T_a\left(f_{x_1} * f_{x_2}\right) \forall x_1, x_2 \in \mathbb{R}$$

*Lemma 2.*

$$\frac{x_+^m}{m!} * \frac{x_+^n}{n!} = \frac{x_+^{m+n+1}}{(m+n+1)!} \quad \forall n, m \in \mathbb{N}, \quad \forall x \in \mathbb{R}$$

By applying lemma 1, we obtain:

$$f_Z(x) = \left(\prod_{j=1}^{d} \frac{1}{2a_j}\right) \times \left(\left(T_{a_1} - T_{-a_1}\right) \cdots \left(T_{a_d} - T_{-a_d}\right) \underbrace{\left(x_+^0\right) * \cdots * \left(x_+^0\right)}_{d \text{ fois}}\right).$$

Then by applying lemma 2:

$$f_Z(x) = \left(\prod_{j=1}^{d} \frac{1}{2a_j}\right) \times \left((T_{a_1} - T_{-a_1})(T_{a_2} - T_{-a_2})\cdots(T_{a_d} - T_{-a_d})\frac{x_+^{d-1}}{(d-1)!}\right).$$

and

$$f_Z(x) = \left(\prod_{j=1}^{d} \frac{1}{2a_j}\right) \times \left(\sum_{s \in \{-1,1\}^d} s_1 \cdots s_d \left(T_{s_1 a_1} \cdots T_{s_d a_d}\right)\frac{x_+^{d-1}}{(d-1)!}\right).$$

Finally,

$$f_Z(x) = \left(\prod_{j=1}^{d} \frac{1}{2a_j}\right) \times \left(\sum_{s} \varepsilon(s)\frac{(x + s.a)_+^{d-1}}{(d-1)!}\right).$$

The announced result is obtained by integrating this relation □

### *Continuity and differentiability of the radar.*

For the sake of simplicity, we give only the proofs in 2 dimensions. However, the result is also valid in 3 dimensions. Let $D^U(z_1,\ldots,z_N) = \sup_{z \in [0,1]} \left|\hat{F}_{z_1,\ldots,z_N}(z) - z\right|$, the Kolmogorov-Smirnov statistic for the uniform distribution over $[0,1]$.

*Proposition.*

(i) $D^U$ is continuous.

(ii) $D^U$ admits partial derivatives with respect to all the $x_i, i=1,\ldots,N$ in $(z_1,\ldots,z_N)$ if and only if the supremum of $\left|\hat{F}_{z_1,\ldots,z_N}(z) - z\right|$ is attained at only one $z_j$, $j \in \{1,\ldots,N\}$.

*Proof.* Let

$$D_N(\theta) = D^U\left(F_\theta(x_1 \cos\theta + y_1 \sin\theta),\ldots,F_\theta(x_N \cos\theta + y_N \sin\theta)\right)$$

where $D^U(z_1,\ldots,z_N) = \sup_{z \in [0,1]} \left|\hat{F}_{z_1,\ldots,z_N}(z) - z\right|$ is the Kolmogorov-Smirnov statistic for the uniform distribution over $[0,1]$. By using the expression of $F_\theta$, we prove that $(\theta, p) \mapsto F_\theta(p)$ is continuous (*cf.* formula (2.1)). Thus, the problem reduces to demonstrate the continuity of $(t_1,\ldots,t_N) \mapsto \sup_t \left|\sum_{i=1}^{N} \frac{1}{N} 1_{\{t > t_i\}} - t\right|$ with respect to all variables. If the values of $t_i$ are perturbed by $\eta_i$, the supremum varies at most by the sum of the $\eta_i$, thereby guaranteeing continuity.

For differentiability, the problem can be reduced to the existence of partial derivatives of $D^U$.

- Let us assume that the supremum is attained for only one $z_j$ where $j \in \{1,...,N\}$.

In this case, the partial derivatives related to $z_i$, $i \neq j$, exist and are equal to 0 because any local perturbation in $z_i$ would not change the supremum. Let us now prove the existence of $\frac{\partial D^U}{\partial z_j}$. Let us assume, for example, that: $\sup_{z \in [0,1]} |\hat{F}_N(z) - z| = \hat{F}_N(z_j) - z_j > 0$. We choose $\varepsilon$ such that $\forall i \neq j, |\hat{F}_N(z_j) - z_j| > |\hat{F}_N(z_i) - z_i| + \varepsilon$ et $\hat{F}_N(z_j) - z_j > \varepsilon$. Then :

$$D^U(z_1,...,z_j + \varepsilon,...,z_N) = D^U(z_1,...,z_j) - \varepsilon \text{ et } D^U(z_1,...,z_j - \varepsilon,...,z_N) = D^U(z_1,...,z_j) + \varepsilon$$

Therefore, $\frac{\partial D^U}{\partial z_j}(z_1,...,z_N)$ exists in $(z_1,...,z_N)$ and is equal to -1.

Similarly, if $\sup_{z \in [0,1]} |\hat{F}_N(z) - z| = z_j - \hat{F}_N(z_j)$, we prove that $\frac{\partial D^U}{\partial z_j}$ exists and is equal to 1.

- Let us assume that the supremum is attained at $z_j$ and at least at another $z_k$ for $j,k \in \{1,...,N\}$. Now, we are going to prove that the partial derivative in $z_j$ does not exist. Assume, for example, that $\hat{F}_N(z_j) - z_j > 0$ and $\varepsilon > 0$, $\hat{F}_N(z_j) - z_j > \varepsilon$. Then we have the following case.

  - The supremum is at $z_j$ and $z_k$, $j \neq k$ where $z_j = z_k$.

Then $D^U(z_1,...,z_j + \varepsilon,...,z_N) = D^U(z_1,...,z_N) - \varepsilon$,

but $D^U(z_1,...,z_j - \varepsilon,...,z_N) = D^U(z_1,...,z_N)$. Therefore, the right-hand derivative is different from the left-hand derivative.

- Otherwise, for all $z_k$, where $k \neq j$ at which the supremum is attained, $z_j \neq z_k$. Thus

$$D^U(z_1,...,z_j - \varepsilon,...,z_N) = D^U(z_1,...,z_N) + \varepsilon$$

but $D^U(z_1,...,z_j + \varepsilon,...,z_N) = D^U(z_1,...,z_N)$ because the local perturbation in $z_j$ does not change the value of the supremum, which is still attained at $z_k$. , the right-hand derivative is therefore also different from the left-hand derivative. □

**Table of the statistic of $G_N$ : values of $g_N$ such that $P = P(G_N < g_N)$.**

| N  | P=0,80 | P=0,85 | P=0,90 | P=0,95 | P=0,99 |
|----|--------|--------|--------|--------|--------|
| 1  | 1,94   | 1,97   | 1,98   | 1,99   | 1,99   |
| 2  | 2,52   | 2,82   | 3,08   | 3,33   | 3,47   |
| 3  | 2,68   | 3,00   | 3,30   | 3,63   | 3,84   |
| 4  | 2,80   | 3,12   | 3,42   | 3,80   | 4,07   |
| 5  | 2,86   | 3,19   | 3,50   | 3,88   | 4,12   |
| 6  | 2,89   | 3,25   | 3,58   | 3,97   | 4,26   |
| 7  | 2,93   | 3,28   | 3,59   | 4,01   | 4,28   |
| 8  | 2,94   | 3,29   | 3,64   | 4,03   | 4,31   |
| 9  | 2,95   | 3,30   | 3,61   | 4,00   | 4,35   |
| 10 | 2,97   | 3,34   | 3,66   | 4,06   | 4,33   |
| 11 | 2,98   | 3,34   | 3,68   | 4,13   | 4,46   |
| 12 | 2,99   | 3,33   | 3,67   | 4,08   | 4,34   |
| 13 | 3,00   | 3,36   | 3,70   | 4,11   | 4,40   |
| 14 | 3,00   | 3,36   | 3,72   | 4,11   | 4,41   |
| 15 | 3,00   | 3,34   | 3,68   | 4,12   | 4,43   |
| 16 | 3,00   | 3,36   | 3,70   | 4,10   | 4,43   |
| 17 | 3,03   | 3,38   | 3,70   | 4,13   | 4,49   |
| 18 | 3,01   | 3,35   | 3,68   | 4,07   | 4,39   |
| 19 | 3,04   | 3,41   | 3,75   | 4,15   | 4,50   |
| 20 | 3,04   | 3,41   | 3,74   | 4,15   | 4,47   |
| 21 | 3,04   | 3,40   | 3,73   | 4,17   | 4,48   |
| 22 | 3,05   | 3,41   | 3,74   | 4,12   | 4,42   |
| 23 | 3,05   | 3,39   | 3,73   | 4,16   | 4,47   |
| 24 | 3,06   | 3,42   | 3,73   | 4,16   | 4,42   |
| 25 | 3,05   | 3,40   | 3,72   | 4,15   | 4,48   |
| 26 | 3,06   | 3,43   | 3,74   | 4,11   | 4,46   |
| 27 | 3,05   | 3,40   | 3,72   | 4,11   | 4,44   |
| 28 | 3,06   | 3,41   | 3,76   | 4,18   | 4,49   |
| 29 | 3,06   | 3,43   | 3,75   | 4,17   | 4,47   |
| 30 | 3,05   | 3,42   | 3,76   | 4,18   | 4,49   |
| 31 | 3,06   | 3,40   | 3,72   | 4,11   | 4,41   |
| 32 | 3,07   | 3,44   | 3,79   | 4,27   | 4,58   |
| 33 | 3,06   | 3,40   | 3,71   | 4,12   | 4,43   |
| 34 | 3,05   | 3,40   | 3,71   | 4,13   | 4,41   |
| 35 | 3,07   | 3,42   | 3,76   | 4,17   | 4,42   |
| 36 | 3,08   | 3,43   | 3,76   | 4,16   | 4,47   |
| 37 | 3,09   | 3,46   | 3,78   | 4,21   | 4,53   |
| 38 | 3,08   | 3,45   | 3,78   | 4,16   | 4,47   |
| 39 | 3,07   | 3,44   | 3,77   | 4,23   | 4,53   |
| 40 | 3,06   | 3,42   | 3,75   | 4,17   | 4,45   |
| 41 | 3,08   | 3,43   | 3,76   | 4,16   | 4,53   |
| 42 | 3,09 | 3,43 | 3,76 | 4,18 | 4,47 |
| 43 | 3,08 | 3,44 | 3,75 | 4,21 | 4,53 |
| 44 | 3,10 | 3,45 | 3,76 | 4,15 | 4,51 |
| 45 | 3,09 | 3,44 | 3,77 | 4,17 | 4,44 |
| 46 | 3,07 | 3,43 | 3,75 | 4,15 | 4,52 |
| 47 | 3,07 | 3,43 | 3,78 | 4,22 | 4,51 |
| 48 | 3,07 | 3,42 | 3,75 | 4,15 | 4,41 |
| 49 | 3,10 | 3,44 | 3,78 | 4,21 | 4,50 |
| 50 | 3,08 | 3,44 | 3,77 | 4,21 | 4,53 |
| 51 | 3,09 | 3,43 | 3,77 | 4,18 | 4,47 |
| 52 | 3,11 | 3,45 | 3,79 | 4,22 | 4,54 |
| 53 | 3,09 | 3,46 | 3,78 | 4,15 | 4,48 |
| 54 | 3,10 | 3,45 | 3,81 | 4,24 | 4,54 |
| 55 | 3,09 | 3,44 | 3,76 | 4,20 | 4,50 |
| 56 | 3,10 | 3,44 | 3,79 | 4,19 | 4,52 |
| 57 | 3,09 | 3,46 | 3,79 | 4,19 | 4,48 |
| 58 | 3,09 | 3,45 | 3,78 | 4,20 | 4,49 |
| 59 | 3,11 | 3,44 | 3,78 | 4,18 | 4,52 |
| 60 | 3,08 | 3,44 | 3,76 | 4,17 | 4,45 |
| 61 | 3,11 | 3,47 | 3,82 | 4,20 | 4,54 |
| 62 | 3,10 | 3,47 | 3,80 | 4,19 | 4,48 |
| 63 | 3,09 | 3,45 | 3,78 | 4,19 | 4,46 |
| 64 | 3,09 | 3,47 | 3,79 | 4,23 | 4,53 |
| 65 | 3,10 | 3,47 | 3,80 | 4,22 | 4,54 |
| 66 | 3,11 | 3,46 | 3,78 | 4,16 | 4,45 |
| 67 | 3,10 | 3,45 | 3,79 | 4,23 | 4,52 |
| 68 | 3,09 | 3,43 | 3,77 | 4,15 | 4,44 |
| 69 | 3,09 | 3,44 | 3,77 | 4,18 | 4,44 |
| 70 | 3,10 | 3,46 | 3,78 | 4,21 | 4,50 |
| 71 | 3,11 | 3,46 | 3,77 | 4,17 | 4,49 |
| 72 | 3,10 | 3,44 | 3,78 | 4,18 | 4,42 |
| 73 | 3,09 | 3,45 | 3,79 | 4,18 | 4,54 |
| 74 | 3,11 | 3,47 | 3,80 | 4,17 | 4,46 |
| 75 | 3,10 | 3,44 | 3,77 | 4,21 | 4,47 |
| 76 | 3,11 | 3,48 | 3,80 | 4,22 | 4,51 |
| 77 | 3,09 | 3,43 | 3,78 | 4,19 | 4,44 |
| 78 | 3,11 | 3,45 | 3,78 | 4,21 | 4,48 |
| 79 | 3,08 | 3,44 | 3,78 | 4,20 | 4,46 |
| 80 | 3,10 | 3,46 | 3,80 | 4,21 | 4,52 |
| 81 | 3,11 | 3,48 | 3,79 | 4,19 | 4,52 |
| 82 | 3,11 | 3,49 | 3,83 | 4,25 | 4,51 |
| 83 | 3,11 | 3,46 | 3,78 | 4,17 | 4,44 |
| 84 | 3,11 | 3,45 | 3,77 | 4,18 | 4,51 |
| 85 | 3,09 | 3,46 | 3,77 | 4,23 | 4,55 |
| 86 | 3,10 | 3,46 | 3,78 | 4,18 | 4,49 |
| 87 | 3,08 | 3,44 | 3,76 | 4,15 | 4,44 |
| 88 | 3,11 | 3,47 | 3,77 | 4,21 | 4,48 |
| 89 | 3,10 | 3,47 | 3,79 | 4,17 | 4,46 |
| 90 | 3,08 | 3,43 | 3,74 | 4,15 | 4,44 |
| 91 | 3,07 | 3,43 | 3,76 | 4,14 | 4,43 |
| 92 | 3,09 | 3,46 | 3,78 | 4,19 | 4,52 |
| 93 | 3,09 | 3,46 | 3,81 | 4,23 | 4,53 |
| 94 | 3,08 | 3,43 | 3,74 | 4,18 | 4,52 |
| 95 | 3,09 | 3,44 | 3,77 | 4,21 | 4,54 |
| 96 | 3,11 | 3,45 | 3,76 | 4,19 | 4,52 |
| 97 | 3,12 | 3,45 | 3,78 | 4,20 | 4,47 |
| 98 | 3,08 | 3,44 | 3,79 | 4,21 | 4,46 |
| 99 | 3,09 | 3,45 | 3,77 | 4,21 | 4,50 |
| 100 | 3,10 | 3,44 | 3,79 | 4,23 | 4,51 |